\def\year{2022}\relax
\documentclass[letterpaper]{article} 
\usepackage{aaai22}  
\usepackage{times}  
\usepackage{helvet}  
\usepackage{courier}  
\usepackage[hyphens]{url}  
\usepackage[table]{xcolor}
\usepackage{graphicx} 
\usepackage{natbib}  
\usepackage{caption} 
\DeclareCaptionStyle{ruled}{labelfont=normalfont,labelsep=colon,strut=off} 
\usepackage{algorithm}
\usepackage{algorithmic}
\usepackage{array}
\usepackage{amsmath}

\usepackage{booktabs}
\usepackage{xcolor}
\usepackage{url}
\usepackage{newfloat}
\usepackage{listings}
\floatstyle{ruled}
\newfloat{listing}{tb}{lst}{}
\floatname{listing}{Listing}
\usepackage{amsthm}

\title{Server-Side Local Gradient Averaging and Learning Rate Acceleration for Scalable Split Learning}
\author {
    $^*$Shraman Pal\textsuperscript{\rm 1}, 
    $^*$Mansi Uniyal\textsuperscript{\rm 1}, 
    Jihong Park \textsuperscript{\rm 2,4},
    Praneeth Vepakomma \textsuperscript{\rm 3},\\
    Ramesh Raskar \textsuperscript{\rm 3},
    Mehdi Bennis \textsuperscript{\rm 5},
    Moongu Jeon \textsuperscript{\rm 4},
    Jinho Choi \textsuperscript{\rm 2}
}
\affiliations {
    \textsuperscript{\rm 1}IIT Kharagpur, India,
    \textsuperscript{\rm 2}Deakin University, Australia,
    \textsuperscript{\rm 3}MIT Media Lab, USA,
    \textsuperscript{\rm 4}GIST, Korea,
    \textsuperscript{\rm 4}University of Oulu, Finland
    \\
    shramanpal@iitkgp.ac.in,
    mansinumber1@iitkgp.ac.in, 
    jihong.park@\{deakin.edu.au, gist.ac.kr\},
    vepakom@mit.edu,
    raskar@mit.edu,
    mehdi.bennis@oulu.fi,
    mgjeon@gist.ac.kr,
    jinho.choi@deakin.edu.au\\
    $^*$Equal contribution
}

\usepackage{bibentry}

\begin{document}

\maketitle

\begin{abstract}
In recent years, there have been great advances in the field of decentralized learning with private data. Federated learning (FL) and split learning (SL) are two spearheads possessing their pros and cons, and are suited for many user clients and large models, respectively. To enjoy both benefits, hybrid approaches such as SplitFed have emerged of late, yet their fundamentals have still been illusive. In this work, we first identify the fundamental bottlenecks of SL, and thereby propose a scalable SL framework, coined SGLR. The server under SGLR broadcasts a common gradient averaged at the split-layer, emulating FL without any additional communication across clients as opposed to SplitFed. Meanwhile, SGLR splits the learning rate into its server-side and client-side rates, and separately adjusts them to support many clients in parallel. Simulation results corroborate that SGLR achieves higher accuracy than other baseline SL methods including SplitFed, which is even on par with FL consuming higher energy and communication costs. As a secondary result, we observe greater reduction in leakage of sensitive information via mutual information using SLGR over the baselines.
\end{abstract}

\begin{figure*}
    \centering
    \includegraphics[width=.9\textwidth]{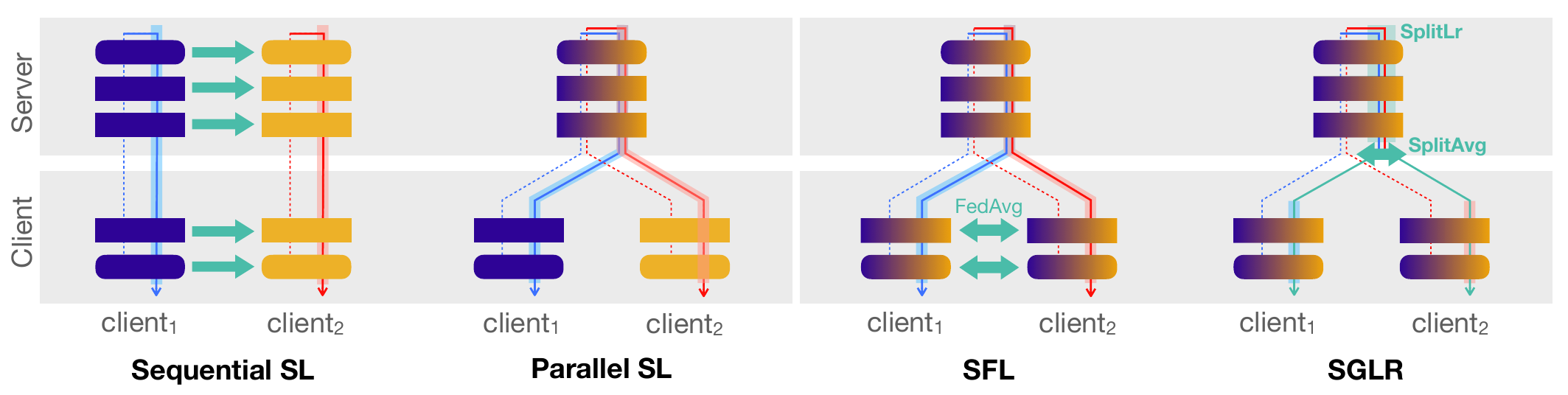}
    \caption{A schematic illustration of sequential split learning (SSL), parallel split learning (PSL), split federated learning (SFL), and the proposed \emph{SGLR with a split-layer gradient averaging (SplitAvg) and learning rate splitting (SplitLr)}, where the shaded area widths represent learning rates, and the dotted solid lines imply forward and backward propagation flows, respectively.}
    \label{fig:overview}
\end{figure*}

\section{Introduction}

The recent trend in deep learning has seen exponential growth in terms of architecture sizes \citet{alom2018history}. In the computer vision domain, the model sizes over the years have grown larger, as observed in the transition from ResNet and VGG \cite{he2015deep,simonyan2015deep} to Inception and DenseNet \cite{szegedy2015rethinking,huang2018densely}. In the field of natural language processing the growth is even more drastic; starting from BERT, RoBERTa, and XLM \cite{devlin2019bert,liu2019roberta,goyal2021largerscale} that have crossed the 100 million parameter mark; and finally reaching Open AI's recent GPT-3 \cite{brown2020language} standing at staggering 175 billion parameters. The prerequisites for running these large models are huge training data and computing power, making them still limited to a select few.

Fortunately for such large models, the current trend in data volume and computing power keeps exponentially increasing in both. Nonetheless, the majority of these data and computing power are sourced from mobile edge devices like smartphones, IoT, and e-health wearables, so are inherently dispersed and often privacy-sensitive \cite{Park2019,smith2018dont}. Towards exploiting the indispensable private data and mobile computing power, recent advances in distributed learning have opened new avenues notably in the form of \emph{federated learning (FL)} \cite{McMahan2016FederatedLO} and \emph{split learning (SL)} \cite{vepakomma2018split}.

In FL, edge devices or \emph{clients} independently train their local models, while periodically exchanging and averaging their local model parameters through a parameter server \cite{McMahan2016FederatedLO}. Consequently, keeping data private, FL can leverage the parallel computing power and global data of clients. However, FL right off the bat removes the chance of using large models, as edge devices cannot store and communicate them due to the strain on memory, bandwidth, and energy \cite{pieee21park}. Alternatively, SL copes with large models by splitting the models into a shared upper segment and distributed lower segments that are respectively stored at the server and clients, respectively. At the \emph{split layer} or \emph{cut layer}, in the forward propagation, each client uploads its final hidden representations, so-called \emph{smashed data}, and downloads its gradient from the server in the backward propagation, thereby keeping local data private.

Notwithstanding, SL has difficulty in achieving scalability, in contrast with FL where the accuracy increases with the number of clients \cite{zhang2021faithful}. As shown by Fig. \ref{fig:overview}, the original SL algorithm, vanilla SL or hereafter referred to as \emph{sequential SL (SSL)}, limits the server to supporting clients one by one after each local training completes. Therefore, SSL requires tight pipelining across clients and fails to leverage parallel computing power. By simply allowing multiple client associations per batch, SSL can easily be extended to its parallel version, termed \emph{parallel SL (PSL)}, which however entails a fundamental issue on the \emph{server-client update imbalance} as elaborated in the following two problems.
\begin{enumerate}
    \item \textbf{Server-Side Large Effective Batch Problem}: The server model is updated multiple times per each client model update, resulting in a larger \emph{effective batch size} (i.e., batch size multiplied by the number of clients, when each client has the same batch size) at the server than that of each client (i.e., batch size).

    \item \textbf{Backward Client Decoupling Problem}: Other clients’ impacts are only reflected in the forward propagation flows when smashed data pass through the server model, whereas the backward propagation flows across clients are decoupled with each other as illustrated in Fig.~1.
\end{enumerate}
A recently proposed \emph{SplitFed learning (SFL)} algorithm addresses the client decoupling problem by additionally applying FL across client models \cite{thapa2021splitfed}. However, the effectiveness of SFL comes at the cost of large communication overhead, and what is more the server-side effective batch size problem remains unsolved.

To fill the aforementioned void, in this article we propose a novel PSL framework, coined \emph{split learning with a gradient averaging and learning rate splitting (SGLR)}. At its core, SGLR aims to address the server-side effective batch size problem, by separating the learning rate into two parts, and accelerating the server model's learning rate, i.e., \emph{learning rate splitting (SplitLr)}, inspired by the techniques for large batch training \cite{smith2018dont}, \cite{goyal2018accurate}. Next, to address the client decoupling problem, SGLR broadcasts the common gradient averaged at the cut-layer from the server to all clients, i.e., \emph{split-layer gradient averaging (SplitAvg)}, rather than unicasting a unique gradient to each client. As opposed to SFL, the gradient averaging of SplitAvg is taken at the server, not incurring any additional communication cost. Even compared to PSL that relies on gradient unicasting through orthogonal bandwidth allocations, SplitAvg leverages broadcasting over the entire bandwidth, making SGLR more communication efficient and scalable particularly under limited bandwidth.

\subsubsection{Contributions \& Organization} The major contributions of this work are outlined as follows.
\begin{itemize}
    \item Motivated by the server-side large effective batch problem, we propose \emph{SplitLr} so as to separately accelerate the learning rate at the server (see \textbf{Sec.~3.1}).
    
    \item Inspired by SFL while identifying its non-negligible communication overhead and possible information leakage incurred by model exchanging and averaging across clients, we develop \emph{SplitAvg} in which the server broadcasts (or multicasts) averaged gradients to clients without incurring any additional communication among clients (see \textbf{Sec.~3.2}).

    \item Combining SplitLr and SplitAvg, we finally propose \emph{SGLR} (see \textbf{Algorithm 1}). Simulation results corroborate the effectiveness of SGLR in terms of accuracy scalability, information leakage, and communication efficiency (see \textbf{Sec. 4.2}, \textbf{4.3}, and \textbf{4.4}, respectively).

\end{itemize}

The rest of this article is organized as follows. In Sec.~2, existing works and their weight/gradient update rules are summarized to identify the server-client weight imbalance problem. In Sec.~3, the operations of SLR are described by elaborating the procedures of SplitAvg and SplitLr algorithms. In Sec.~4, the effectiveness of SGLR is validated by simulation and comparison with other baseline frameworks such as \emph{centralized learning (CL)}, FL, sequential/parallel SL, and SFL. Finally, we conclude this article by discussing several future research directions in Sec.~5.


\section{Preliminaries: FL, SL, and SFL}

Federated learning \cite{McMahan2016FederatedLO} and Split learning \cite{gupta2018distributed}, \cite{vepakomma2018split} are two new frameworks that allow training a model effectively from various distributed data sources without sharing the raw data. The device that possesses the data is usually termed as the \emph{client} while the generally computationally powerful device is termed as the \emph{server} which could have multiple functions based on the framework considered. In all the frameworks, there exist a set of clients $\mathcal{C}$, where each \emph{i}-th client has their data $\mathcal{D}_i$ and model weights $\mathbf{w_i}$ stored locally. $\mathcal{D} = \bigcup_{i \in \mathcal{C}}\mathcal{D}_i$ represents the total data. 

We represent the model weights of the \emph{i}-th client as, $\mathbf{w}_{c,i} = [w_{c,i}^{L_c}, w_{c,i}^{L_c-1},...,w_{c,i}^{1}]^T$ where $L_c$ is the final layer in the client model. The shared server model for SL frameworks is denoted by $\mathbf{w}_{s} = [w_{s}^{L}, w_{s}^{L-1},...,w_{s}^{L_c+1}]^T$ where $L$ is the last layer of the server model. In the next subsections, we briefly outline the frameworks, which will helps us highlight the benefits and limitations of each method on different aspects of latency, model performance and communication efficiency.

\subsection{Sequential Split Learning (SSL)}
In SSL which is also called vanilla split learning \cite{vepakomma2018split}, \cite{gupta2018distributed}, a lower model segment, not necessarily the same, is present in multiple clients and an upper model segment is present on a shared server. 
During training, each client is sequentially selected from the set of all clients. This selected \emph{i}-th client selects a mini batch $\mathcal{B}_i \in \mathcal{D}_i$ which consists of \emph{b} input-label tuples $(x_{i,j}, y_{i,j})$ from its local data. The \emph{i}-th client produces \emph{b} smashed data $s_{i,j} = f(\mathbf{w}_{c,i}, x_{i,j})$ by passing the \emph{j}-th input data $x_{i,j}$ through it for all $j \in B_i$.

The client uploads the smashed data-label tuples to the server which produces the final predictions $y'_{i,j} = f(\mathbf{w}_{s}, s_{i,j})$. The loss for the batch is denoted as $L(\mathbf{w}_{c,i}, w_s)$  where the $L$ is the required loss function. The optimisation objective therefore simply becomes $min_{\mathbf{w}_{c,i}, \mathbf{w}_s}\{L(\mathbf{w}_{c,i}, \mathbf{w}_s)\}$. 

\subsubsection{Gradients}
Using the evaluated loss, the server calculates the gradients and backpropagates(BP) for its model layers:
\begin{align}
    \mathbf{g}_{s}^{\text{SSL}} = \begin{bmatrix}
    g_{s}^{L}\\
    g_{s}^{L-1}\\
    ...\\
    g_{s}^{L_c+1}\\
    \end{bmatrix}
    =
    \begin{bmatrix}
    \nabla_{\mathbf{w}_{s}^L}L(\mathbf{w}_s,\mathbf{w}_{c,i})\\
    \text{BP}(g_{s}^{L}) \\
    ... \\
    \text{BP}(g_{s}^{L_c+2}) \\
    \end{bmatrix}.
\end{align}
The server sends the gradient of the \emph{cut\_layer} $g_{s}^{L_c+1}$ to the \emph{i}-th client. Then the \emph{i}-th client generates the gradient for its own model segment:
\begin{align}
    \mathbf{g}_{c,i}^{\text{SSL}} = \begin{bmatrix}
    g_{c,i}^{L_c}\\
    g_{c,i}^{L_c-1}\\
    ...\\
    g_{c,i}^{1}\\
    \end{bmatrix}
    =
    \begin{bmatrix}
    \text{BP}(g_{s}^{L_c+1})\\
    \text{BP}(g_{c,i}^{L_c}) \\
    ... \\
    \text{BP}(g_{c,i}^{2}) \\
    \end{bmatrix}.
\end{align}

\subsubsection{Weight Updates}
The weight updates for both the client and the server thereafter can be written in a compact form as:
\begin{align}
\label{weight update ssl}
\begin{bmatrix}
\mathbf{w}_{s}\\
\mathbf{w}_{c,i}
\end{bmatrix}
\leftarrow
\begin{bmatrix}
\mathbf{w}_{s}\\
\mathbf{w}_{c,i}
\end{bmatrix}
-\eta
\begin{bmatrix}
\mathbf{g}_{s}^{\text{SSL}}\\
\mathbf{g}_{c,i}^{\text{SSL}}
\end{bmatrix}.
\end{align}
After that the lower model segment $\mathbf{w}_{c,i}$ is sent to the (\emph{i}+1)-th client $\mathbf{w}_{c,i+1} \leftarrow \mathbf{w}_{c,i}$ where $(i+1) \in \mathcal{C}\setminus\{i\}$. Thereafter the \emph{i}+1-th client continues with its local iterations.

\subsubsection{Benefits and Limitations}
SSL can achieve high accuracy due to the shared server being able to learn from multiple local datasets which make the weights robust. Also due to the lower model weight sharing, there is information sharing between the clients directly as well. A smaller lower model segment allows the use of edge devices. The latency, however, increases linearly with the increase in the number of clients which is a huge drawback as it fails to capture the potential of parallel computing which motivates the use of PSL.

\subsection{Parallel Split Learning (PSL)}
In PSL, the server and clients are the same as in SSL but all clients are connected to the server simultaneously. All the clients run their forward passes in parallel and upload their smashed data to the server together. The server generates a loss $L(\mathbf{w}_{c,i}, \mathbf{w}_s)$ corresponding to each client's smashed data in parallel.

\subsubsection{Gradients}
Then the server computes the gradients for its layers by weighing the individual losses according to the size of the data of each client. The gradient for each layer of the server is cast as:
\begin{align}
\label{server grad psl}
    \mathbf{g}_{s}^{\text{PSL}} = \Sigma_{i \in \mathcal{C}}\delta_i\begin{bmatrix}
    g_{s,i}^{L}\\
    g_{s,i}^{L-1}\\
    \cdots \\
    g_{s,i}^{L_c+1} \\
    \end{bmatrix}
    = 
    \Sigma_{i \in \mathcal{C}}\delta_i\begin{bmatrix}
    g_{s,i}^{L}\\
    \text{BP}(g_{s,i}^{L}) \\
    \cdots \\
    \text{BP}(g_{s,i}^{L_c+2}) \\
    \end{bmatrix},
\end{align}
where $\delta_i=\frac{|\mathcal{D}_i|}{|\mathcal{D}|}$ . The local gradients are sent to the corresponding clients with which the clients calculate the gradients for the layers of its own model. The client-side gradient is given as:
\begin{align}
\label{client grad psl}
    \mathbf{g}_{c,i}^{\text{PSL}} = \begin{bmatrix}
    g_{c,i}^{L_c}\\
    g_{c,i}^{L_c-1} \\
    ...  \\
    g_{c,i}^{1} \\
    \end{bmatrix}
    = \begin{bmatrix}
    \text{BP}(g_{s,i}^{L_c+1})\\
    \text{BP}(g_{c,i}^{L_c}) \\
    ...\\
    \text{BP}(g_{c,i}^{2}) \\
    \end{bmatrix}.
\end{align}
Note that while the feed-forward flows propagate in parallel, the backward flows propagate sequentially.

\subsubsection{Weight Updates}
\eqref{weight update ssl}
The weight update for PSL remains the same as \eqref{weight update ssl} of SSL, i,e.,
\begin{align}
\label{weight update psl}
\begin{bmatrix}
\mathbf{w}_{s}\\
\mathbf{w}_{c,i}
\end{bmatrix}
\leftarrow
\begin{bmatrix}
\mathbf{w}_{s}\\
\mathbf{w}_{c,i}
\end{bmatrix}
-\eta
\begin{bmatrix}
\mathbf{g}_{s}^{\text{PSL}}\\
\mathbf{g}_{c,i}^{\text{PSL}}
\end{bmatrix}.
\end{align}
It is worth noting that SSL is a special case of PSL where $\delta_i=0$ and the rest are $0$. In this case only the \emph{i}-th client is given importance for the backward pass and essentially neglecting the rest.

\subsubsection{Benefits and Limitations}
PSL achieves lower latency than SSL as the forward pass happens in parallel utilising the parallel computing of edge devices to the utmost. However, the performance falls below SSL due to the server-client update imbalance problem. The client's local gradient is dependent only on a single batch of input data while, unlike SSL, the server's gradient depends on multiple batches. The multiple batches at the server also lead to a high effective batch size. Also as the clients only utilise their local gradient, the clients get \emph{detached} during the backward pass, which we term as the \emph{backward client decoupling} problem. We elaborate on these in the later sections.

\subsection{Federated Learning (FL)}
In FL or specifically \emph{FedAvg}, every client runs an exact copy of the entire model on its own local data where each client~\emph{i} focuses on the local optimization task $\min_{w_{c,i}}\{L(\mathbf{w_{c,i}})$. The server in this framework only acts as an aggregator and stores no models weights. It receives the local model weights from each client and takes a weighted mean (\emph{FedAvg}) to get the \emph{global weights}. These global weights which intuitively are more generalized are downloaded by every client. The clients resume training using these global weights. This process continues till convergence. We only take the algorithm of \emph{FedAvg} out of several existing algorithms as it is the closest comparison to ours in terms of how it works.

\subsubsection{Gradients}
FL does not consider model-split architectures, and the server does not train any models. Therefore there are no gradients at the server side. The gradients at the client side are given as
\begin{align}
    \mathbf{g}_{c,i}^{\text{FL}} = \begin{bmatrix}
    g_{c,i}^{L}\\
    g_{c,i}^{L-1} \\
    \cdots  \\
    g_{c,i}^{1}\\
    \end{bmatrix}
    = \begin{bmatrix}
    \nabla_{w_{c,i}^L}L(w_{c,i})\\
    \text{BP}(g_{c,i}^{L}) \\
     \cdots \\
    \text{BP}(g_{c,i}^{2}) \\
    \end{bmatrix}.
\end{align}
Note again that each client under FL stores $L$ layers, i.e., an entire model. This is in contrast to the client under SL storing $L_c$ layers while the server stores the remaining $L-L_c$ layers.

\subsubsection{Weight Updates}
The weight update using these gradients can be written in a compact form as:
\begin{align}
\mathbf{w}_{c,i} \leftarrow \Sigma_{k \in \mathcal{C}} \delta_k \{ \mathbf{w}_{c,k} - \eta \mathbf{g}_{c,k}^{\text{FL}} \}.
\label{weight update fl}    
\end{align}
In other words, once each client has updated its local weights $\mathbf{w}_{c,i}$, they upload it to the server which averages them. Afterwards, the clients download the averaged weights and assigns them to their local weights 
for the next iteration.

\subsubsection{Benefits and Limitations}
FL achieves high accuracy, comparable to CL while maintaining a low latency due to its parallel computing clients. With small models, the communication overhead is low as well. However, with larger models, not only would we require powerful devices with large storage capacity but it would also result in large communication overhead. Such computationally powerful devices are infeasible when we consider edge devices as clients.

\subsection{Split Federated Learning (SFL)}
SFL \cite{thapa2021splitfed} combine FL and SL by averaging the model weights after updating the weights similar to FL but only limited to the lower model segments stored on the clients. Local Model Weight Averaging or \emph{LocAvg} is used to solve the backward client decoupling problem. The forward pass, the loss $L(\mathbf{w}_{c, i}, \mathbf{w}_s)$ computation at the server-side, and the backward pass remains the same as PSL. The extra step is to average the client model weights in a layer-wise fashion.

\subsubsection{Gradients}
Similar to PSL, the server gradient remains the same $g_{s}^{\text{SFL}} = g_{s}^{\text{PSL}}$. The client gradients can be written equivalently as a weighted sum of all the gradients instead of a weighted sum of the weights. Accordingly, the client gradients are cast as:
\begin{align}
    \mathbf{g}_{c,i}^{\text{SFL}} = \Sigma_{k \in \mathcal{C}}\delta_k\begin{bmatrix}
    g_{c,k}^{L_c}\\
    g_{c,k}^{L_c-1}\\
    ...\\
    g_{c,k}^{1}\\
    \end{bmatrix}
    =
    \Sigma_{k \in \mathcal{C}}\delta_k\begin{bmatrix}
    \text{BP}(g_{s,k}^{L_c+1})\\
    \text{BP}(g_{c,k}^{L_c}) \\
    \cdots \\
    \text{BP}(g_{c,k}^{2}) \\
    \end{bmatrix}.
\end{align}
We emphasize that the actual framework, averages the weights but as long as the weight averaging happens after each client weight update, they can be written as a weight update with averaged gradients.

\subsubsection{Weight Updates}
The weight updates of SFL can be simplified and written in a similar form of \eqref{weight update ssl} as follows:
\begin{align}
\label{weight update sfl}
\begin{bmatrix}
\mathbf{w}_{s}\\
\mathbf{w}_{c,i}
\end{bmatrix}
\leftarrow
\begin{bmatrix}
\mathbf{w}_{s}\\
\mathbf{w}_{c,i}
\end{bmatrix}
-\eta
\begin{bmatrix}
\mathbf{g}_{s}^{\text{SFL}}\\
\mathbf{g}_{c,i}^{\text{SFL}}
\end{bmatrix}.
\end{align}
Note here that the averaging comes after completing the backpropagation across all the layers.

\subsubsection{Benefits and Limitations}
SFL keeps the latency improvement of PSL by running computations in parallel. By employing the idea of FL, specifically \emph{LocAvg}, it solves the backward client decoupling problem by sharing direct information across the clients. However, this leads to a larger communication overhead than either PSL or FL. Not to mention that the large effective batch size problem at the server remains unattended. In the next section, we aim to tackle all the problems using our proposed novel framework.

\section{SGLR: PSL with Split-Layer Gradient Averaging and Learning Rate Splitting}
We utilise ideas from existing learning rate acceleration work \cite{goyal2018accurate}, \cite{krizhevsky2014weird}, to derive a new scaling rule that works on the notion of effective batch size at the server-side and we call it \emph{SplitLr}. We also derive ideas from FL to adopt a new method to average gradients only at the cut layer, called \emph{SplitAvg} to solve the backward pass decoupling problem.

The forward pass at the clients is the same as PSL. As the server concatenates the smashed data along the batch dimension, after receiving it from the clients, it increases the effective batch size at the server which denotes how many samples pass through the server in one forward pass before any weight update takes place. The effective batch size at the server increases almost linearly with the number of clients. The \emph{concatenated} batch at the server denoted by $\mathcal{B}_{s}$ has an effective batch size of
\begin{align}\label{eff server batch}
    |\mathcal{B}_{s}| = \Sigma_{i \in \mathcal{C}}|\mathcal{B}_{i}|,
\end{align}
where $\mathcal{B}_i$ denotes the batch for the \emph{i}-th client as used before.



Following the forward pass through the server,  we obtain one loss value $L(\mathbf{w}_{c,i}, \mathbf{w}_{c,2},...,\mathbf{w}_{c,|\mathcal{C}|},\mathbf{w}_s)$ for all the clients combined unlike PSL which generated a loss value for each client. The single loss value is used to generate the gradients of the server layers and the local gradients of the clients. Next, we sample a subset of clients and term them \emph{active clients} whose local gradients are averaged which we term as \emph{SplitAvg}. This averaged gradient is substituted for the local gradient of the active clients and the remaining clients utilise their corresponding local gradients.

This tackles the backward pass decoupling problem as the averaged gradient now contains information of multiple clients. This allows information to flow from one client to another indirectly without requiring the communication of client model weights. This framework results in a scalable system with less communication overhead than any SL framework. We present the two algorithms \emph{SplitLr} and \emph{SplitAvg} in the next two subsections along with detailed gradient calculation and weight updates.

\subsection{SplitLr: Learning Rate Splitting}
In the existing literature, there are several learning rate schemes based on the batch size \cite{goyal2018accurate}, \cite{krizhevsky2014weird} which outperform the baseline. They show that for optimizers like Stochastic Gradient Descent (SGD) a linear scaling rule $\eta = \eta_0|\mathcal{B_i}|$ is useful whereas for optimizers like Adam a square root scaling rate $\eta = \eta_0(|\mathcal{B_i}|)^{0.5}$ achieves better performance. We do not use different batch sizes at the client side but as shown in \eqref{eff server batch}, the effective batch size in the server is different. Therefore we use the learning rate scaling only for the server-side model.

The utilisation of a different learning rate scheme for a segment of the model is new, which forces us to assume a generic scaling rule and perform experimental simulations to identify the optimal scaling rule. The learning rate scaling rule can be reduced a power-law model:
\begin{align} \label{server variable lr alpha}
    \eta_s = \eta_0|\mathcal{B}_s|^\alpha.
\end{align}
The hyperparameter $\alpha$ is iterated with values ranging from 0.0 to 2.0 with a step size of 0.5. The case of $\alpha=0.0$ is the setting of no SplitLr.

We extend the equation further to include a special case where each client has the same batch size $|\mathcal{B}_i| = b$ which simplifies $|\mathcal{B}_s|=b|\mathcal{C}|$ and the learning rate scheme becomes:
\begin{align}
    \eta_s &= \eta_0(b|\mathcal{C}|)^\alpha= \eta_c(|\mathcal{C}|)^\alpha,
\end{align}
where $\eta_c = \eta_0b^\alpha$ as $b$ is a constant and can be merged into $\eta_0$. This equation shows scaling that is proportional to the number of clients rather than the batch size at the client side. We utilise this equation for our simulation runs. By combining PSL with SplitLr, we name the framework as \emph{SLR}.

The gradient at the server side and client side for SLR is the same as \eqref{server grad psl} and \eqref{client grad psl}. Therefore $\mathbf{g}_{s}^{\text{SLR}}=\mathbf{g}_{s}^{\text{PSL}}$ and $\mathbf{g}_{c,i}^{\text{SLR}}=\mathbf{g}_{c,i}^{\text{PSL}}$. The modification lies in the weight update step where it becomes:
\begin{align}
\label{weight update slr}
\begin{bmatrix}
\mathbf{w}_{s}\\
\mathbf{w}_{c,i}
\end{bmatrix}
\leftarrow
\begin{bmatrix}
\mathbf{w}_{s}\\
\mathbf{w}_{c,i}
\end{bmatrix}
-
\begin{bmatrix}
\eta_s\mathbf{g}_{s}^{\text{SLR}}\\
\eta_c\mathbf{g}_{c,i}^{\text{SLR}}
\end{bmatrix}.
\end{align}
The key point is simply the different learning rate that we employ for the server side and the client side.

\subsection{SplitAvg: Split Layer Gradient Averaging}
Unlike SFL, which averages model weights, we aggregate the local gradients of a subset of clients and term it \emph{SplitAvg}. Averaging the gradients decreases the communication overhead compared to any SL framework. The accuracy also increases as we present in Tab.~\ref{Results Comparison}. The averaging of the local gradients allows the subset of participating clients to share some information in the form of gradients throughout the training which tackles the backward client decoupling problem. By combining PSL with SplitAvg We name this framework as \emph{SGL}.

The gradients for the server remain as PSL in  \eqref{server grad psl} and can be simply stated as $\mathbf{g}_{s}^{\text{SGL}}=\mathbf{g}_{s}^{\text{PSL}}$ but the gradients of the client models are different. For the active clients denoted by $\mathcal{C}_a = \phi\mathcal{C}$ where $\phi$ denotes the fraction of total clients which are active, the client-side gradients are given by:
\begin{align}
    \mathbf{g}_{c,i,a}^{SGL} = \begin{bmatrix}
    g_{c,i,a}^{L_c}\\
    g_{c,i,a}^{L_c-1}\\
    ...\\
    g_{c,i,a}^{1}\\
    \end{bmatrix}
    =
    \begin{bmatrix}
    \text{BP}(\Sigma_{k \in \mathcal{C}_a}g_{s,k}^{L_c+1})\\
    \text{BP}(g_{c,i,a}^{L_c}) \\
    ... \\
    \text{BP}(g_{c,i,a}^{2}) \\
    \end{bmatrix},
\end{align}
where subscript \emph{a} identifies the client activation, and the \emph{i}-th client belongs to the set of active clients $\mathcal{C}_a$. For clients which are not active have the same gradients as in PSL with \eqref{client grad psl}. Their gradients  $\mathbf{g}_{c,i}^{\text{SGL}}=\mathbf{g}_{c,i}^{\text{PSL}}$ as they are equal.

The weight update for the active clients utilise this averaged gradient while the rest use the corresponding local gradients:
\begin{align}
\label{weight update sgl}
\begin{bmatrix}
\mathbf{w}_{s}\\
\mathbf{w}_{c,i,a}\\
\mathbf{w}_{c,i}
\end{bmatrix}
\leftarrow
\begin{bmatrix}
\mathbf{w}_{s}\\
\mathbf{w}_{c,i,a}\\
\mathbf{w}_{c,i}
\end{bmatrix}
-\eta
\begin{bmatrix}
\mathbf{g}_{s}^{\text{SGL}}\\
\mathbf{g}_{c,i,a}^{\text{SGL}}\\
\mathbf{g}_{c,i}^{\text{SGL}}
\end{bmatrix}.
\end{align}
It is worth noting that in SFL the averaging occurs after the total backpropagation has taken place while in SGL the gradient is averaged before the backpropagation at the client-side takes place.

\begin{figure}
    \centering
    \includegraphics[width=\columnwidth]{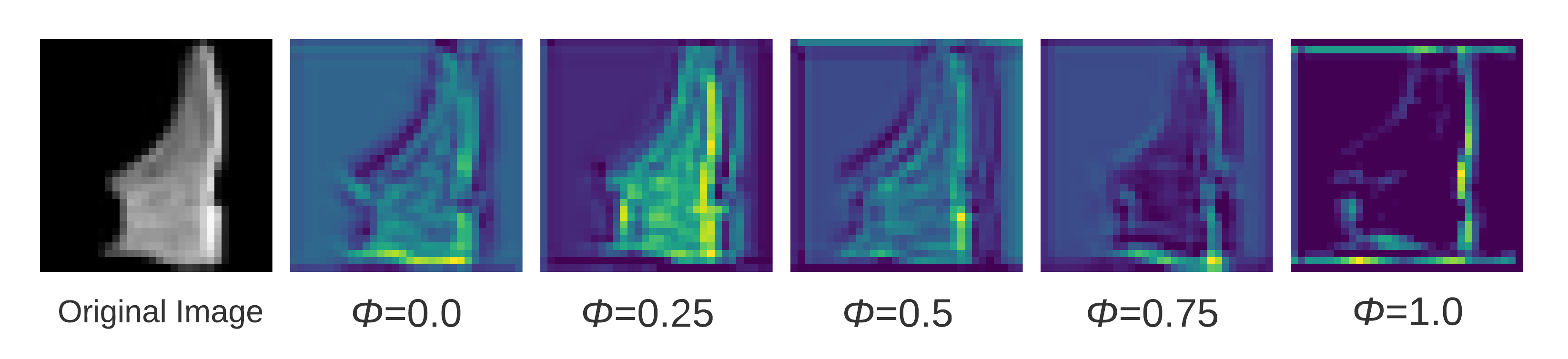}
    \caption{Intermediate Representations of the trained ResNet model. The different variations of using Local Gradient Averaging for constant server learning rate.}
    \label{fig:intermediate activations}
\end{figure}

Taking the fraction $\phi=1.0$ can lead to the training not converging as we show in the results section. Therefore we also use a phased training strategy that utilises the averaged gradient for a certain phase before switching to the local gradients completely or vice-versa. SplitAvg also acts as an inherent regularizer which helps prevent over-fitting. In the Fig.~\ref{fig:intermediate activations} we can observe that as we increase the fraction of clients which participate in the local gradient averaging, the edges or the activation of the image become less sharp i.e the model becomes more robust capable of classifying unseen images better. 

\subsection{SGLR: PSL with SplitLr \& SplitAvg}
The two algorithms combined give the final framework SGLR. The SplitLr method addresses the large effective batch size problem while the SplitAvg addresses the backward decoupling problem. The final weight update by combining both the methods is given as:
\begin{align}
\label{weight update sglr}
\begin{bmatrix}
\mathbf{w}_{s}\\
\mathbf{w}_{c,i,a}\\
\mathbf{w}_{c,i}
\end{bmatrix}
\leftarrow
\begin{bmatrix}
\mathbf{w}_{s}\\
\mathbf{w}_{c,i,a}\\
\mathbf{w}_{c,i}
\end{bmatrix}
-
\begin{bmatrix}
\eta_s\mathbf{g}_{s}^{\text{SGL}}\\
\eta_c\mathbf{g}_{c,i,a}^{\text{SGL}}\\
\eta_c\mathbf{g}_{c,i}^{\text{SGL}}
\end{bmatrix}.
\end{align}
Here, the \emph{i}-th client is not active for all iterations.

The crux of our changes has only been in the gradients of the clients and the learning rates schemes at the server. The performance improvement of our method over the other existing works barring FL as shown in Tab.~ \ref{Results Comparison}, highlights the problem of backward decoupling existing in SL. It might be interesting to note that it is quite possible that averaging only a subset of client model weights in SFL might lead to higher performance. The detailed algorithm combining both the methods has been outlined in \ref{alg multicast:cap}.
\begin{algorithm}[t]
    \caption{SGLR: Parallel SL with SplitLr \& SplitAvg}\label{alg multicast:cap}
    \begin{algorithmic}[1]
    \STATE \textbf{SplitLr}
    \STATE $\eta_c = \eta_0$ 
    \COMMENT {Client Lr}
    \STATE $\eta_s = \eta_0C^\alpha$
    \COMMENT {Server Lr}
    
    \item[]
    \FOR{$epoch \gets (0 - E)$}
        \STATE \textbf{/*Runs on clients*/}
        \FOR{each client $i \in \mathcal{C}$ in parallel}
            \STATE $s_i \gets f(\mathbf{w}_{c,i}, \mathcal{B}_i)$
            \STATE Upload: $(s_i, Y_i)$ from $\text{client}_i$ to server
        \ENDFOR
        \item[]
        \STATE \textbf{/*Runs on server*/}
        \STATE $s \gets [s_1,s_2, \cdots,s_{|\mathcal{C}|}]$
        \STATE $Y \gets [Y_1, Y_2,\cdots,Y_{|\mathcal{C}|}]$
         \STATE $Y_p \gets f(\mathbf{w}_s, s)$
        \STATE Calculate loss $L(\mathbf{w}_{c,1},\cdots,\mathbf{w}_{c,|\mathcal{C}|},\mathbf{w}_{s})$
        \STATE Calculate gradient $\mathbf{g}_{s}^{\text{SGLR}}$ 
        \STATE Weight Update $\mathbf{w}_{s} \leftarrow \mathbf{w}_{s} - \eta_s\mathbf{g}_{s}^{\text{SGLR}}$
        
        \STATE \textbf{SplitAvg} 
        \STATE $\mathcal{C}_a \gets \phi\mathcal{C}$
        \COMMENT {Sampled active clients}
        \STATE Download average gradient $\Sigma_{k \in \mathcal{C}_a}\mathbf{g}_{s,k}^{L_c+1}$
        \STATE Download local gradient $\mathbf{g}_{s,k}^{L_c+1}$ for $k \in \mathcal{C}\setminus\mathcal{C}_a$
        \item[]
        \STATE \textbf{/*Runs on clients*/}
        \FOR{each client $i \in \mathcal{C}$ in parallel}
            \IF{$i \in \mathcal{C}_a$}
                \STATE Calculate gradient $\mathbf{g}_{c,i,a}^{\text{SGLR}}$
                \STATE Weight Update $\mathbf{w}_{c,i} \leftarrow \mathbf{w}_{c,i} - \eta_c\mathbf{g}_{c,i,a}^{\text{SGLR}}$ 
            \ELSE
                \STATE Calculate gradient $\mathbf{g}_{c,i}^{\text{SGLR}}$
                \STATE Weight Update $\mathbf{w}_{c,i} \leftarrow \mathbf{w}_{c,i} - \eta_c\mathbf{g}_{c,i}^{\text{SGLR}}$
            \ENDIF
        \ENDFOR
    \ENDFOR
    \end{algorithmic}
\end{algorithm}
The next section presents our experimental results in terms of accuracy, information leakage, and communication efficiency.

\section{Simulation Results}
This section contains the results for the different runs we perform with our architecture. The section validates that our architecture and training strategies are scalable while maintaining a very low computational and memory requirement.

\subsection{Experimental Settings}
The default setting for all the experiments in this section is using a Resnet16 architecture where the first 9 layers are stored on each of the client devices and the last 7 layers are stored on the server. The division has been done keeping in mind that the later layers have a much larger number of channels and require larger memory which is suitable for the server. We conduct our runs on the Fashion MNIST dataset \cite{xiao2017fashionmnist} which contains 60000 training images and 10000 test images.

We separate 10000 images for the validation set from the training set such that it is i.i.d in nature. Each client possesses 1000 i.i.d samples which are randomized on each run. We use Adam \cite{kingma2017adam} with a client learning rate of $10^{-3}$ and default momentum values of $\beta_1 = 0.9$, $\beta_2 = 0.999$ as our optimizer and vary the server learning rate according to the SplitLr setting. We utilise a batch size of 8 taking into consideration that edge devices do not possess much RAM. Having laid down the basic settings of our experimental runs we move to the actual results.

\subsection{Accuracy Scalability}
As the number of clients increases, we present the top-1 accuracy across different settings, where the baseline is when no algorithm, SplitAvg or SplitLr, is used. We look at the individual impact of SplitLr and SplitAvg before observing how combining the two algorithms performs.

\subsubsection{Impact of SplitLr}
The SplitLr algorithm aims to utilise the large effective batch size at the server to its advantage.  The use of SplitLr improves on the results providing higher accuracy across the board leading to nearly a 1.5\% increase for 20 clients. The results for learning rate splitting have been shown in Tab.~\ref{Results 1}.

\begin{table}[ht]
\centering
 \resizebox{.8\columnwidth}{!}{
    \begin{tabular}{c c c c c c}
    \toprule[1pt]
    \textbf{C} & {\color{gray}\textbf{w.o. SplitLr}} & \multicolumn{4}{c}{\textbf{SplitLr}}\\
    & & $\mathbf{\alpha}$\textbf{=2.0} & $\mathbf{\alpha}$\textbf{=1.5} & $\mathbf{\alpha}$\textbf{=1.0} & $\mathbf{\alpha}$\textbf{=0.5} \\
    \cmidrule(lr){1-1} \cmidrule(lr){2-2} \cmidrule(lr){3-6}
    &\\[-1.5ex]
    1 & {\color{gray}82.45} & 82.45 & 82.45 & 82.45 & 82.45 \\
    2 & {\color{gray}82.04} & \textbf{82.11} & 82.00 & 81.68 & 81.76 \\
    4 & {\color{gray}82.68}  & \textbf{82.14} & \textbf{83.42} & 82.93 & 82.46 \\
    6 & {\color{gray}82.84} & 83.56 & 84.21 & \textbf{84.58} & 83.30 \\
    8 & {\color{gray}83.49} & 58.41 & 83.87 & \textbf{84.52} & 84.23 \\
    10 & {\color{gray}83.49} & 34.37 & 84.21 & \textbf{84.98} & 84.78 \\
    15 & {\color{gray}84.34} & - & 84.08 & \textbf{85.79} & 85.30 \\
    20 & {\color{gray}84.88} & - & - & 86.00 & \textbf{86.25} \\ [0.5ex]
    \bottomrule[1pt]
    \end{tabular}
    }
\caption{Top-1 accuracy for different SplitLr settings. Lower values of $\alpha$ perform better as the number of clients increase.}  
\label{Results 1}
\end{table}

As observed in Tab.~\ref{Results 1}, large values of $\alpha$ do well only for the small number of clients while smaller values perform well for a large number of clients. This is quite understandable as \cite{kingma2017adam} states that Adam is quite sensitive to learning rate and too large values often leads to instability. As an example using $\alpha$ close to 2.0, for 20 clients, the learning rate becomes 400 times larger. For a client learning rate of $10^{-3}$, the server learning rate becomes $0.4$ which is very large for Adam and is not used in the existing literature. Therefore for large clients, we suggest using smaller values of $\alpha$ while for a smaller number of clients larger values can be utilised.

\begin{table*}[t]
\centering
    \resizebox{.85\textwidth}{!}{
    \begin{tabular}{ccccccccccccc}
    \toprule[1pt]
    \textbf{\emph{C}} & 
    \multicolumn{4}{c}{\textbf{Constant learning rate}} &  \multicolumn{4}{c}{\textbf{SplitLr} $\mathbf{\alpha=0.5}$} &  \multicolumn{4}{c}{ \textbf{SplitLr} $\mathbf{\alpha=1.0}$}\\[0.5ex]
      & 
     \textbf{{\color{gray}w.o. SplitAvg}} & \multicolumn{3}{c}{\textbf{SplitAvg}} & \textbf{w.o. SplitAvg} & \multicolumn{3}{c}{\textbf{SplitAvg}} & \textbf{w.o. SplitAvg} & \multicolumn{3}{c}{\textbf{SplitAvg}}\\[0.5ex]
     & & 
     \textbf{\emph{f}=0.25} & \textbf{\emph{f}=0.5} & \textbf{\emph{f}=0.75} 
     & &
     \textbf{\emph{f}=0.25} & \textbf{\emph{f}=0.5} & \textbf{\emph{f}=0.75}
     & &
     \textbf{\emph{f}=0.25} & \textbf{\emph{f}=0.5} & \textbf{\emph{f}=0.75}\\
     \cmidrule(lr){1-1}
     \cmidrule(lr){2-2}
     \cmidrule(lr){3-5}
     \cmidrule(lr){6-6}
     \cmidrule(lr){7-9}
     \cmidrule(lr){10-10}
     \cmidrule(lr){11-13}
    1  & {\color{gray}82.45} & 82.45 & 82.45 & 82.45 & 82.45 & 82.45 & 82.45 & 82.45 & 82.45 & 82.45 & 82.45 & 82.45\\
    2  & {\color{gray}82.04} & 81.91 & 81.84 & \textbf{81.94} & 81.76 & 81.88 & \textbf{81.93} & \textbf{81.92} & 81.68 & 81.95 & \textbf{82.10} & 81.72 \\
    4  & {\color{gray}82.68} & 82.90 & 83.16 & \textbf{83.81} & 82.46 & 82.78 & 83.24 & \textbf{84.22} & 82.93 & 83.30 & 83.66 & \textbf{84.45} \\
    6  & {\color{gray}82.84} & 83.20 & 84.02 & \textbf{84.08} & 83.30 & 83.06 & \textbf{84.89} & 84.62 & 84.28 & 83.70 & \textbf{84.32} & 84.24\\
    8  & {\color{gray}83.49} & 83.73 & 84.27 & \textbf{85.14} & 84.23 & 84.33 & 85.04 & \textbf{85.29} & 84.52 & 84.58 & 85.41 & \textbf{85.68}\\
    10 & {\color{gray}83.49} & 84.35 & 84.74 & \textbf{85.49} & 84.78 & 84.65 & 85.22 & \textbf{85.93} & 84.98 & 84.82 & 85.89 & \textbf{86.11} \\
    15 & {\color{gray}84.34} & \textbf{85.87} & \textbf{85.86} & \textbf{85.86} & 85.30 & 85.89 & 86.71 & \textbf{87.22} & 85.79 & 85.19 & 86.15 & \textbf{86.20}\\
    20 & {\color{gray}84.88} & 86.33 & 86.59 & \textbf{86.82} & 86.25 & 86.96 & \textbf{87.48} & \textbf{87.47} & 86.00 & 86.54 & \textbf{87.11} & 86.67\\[0.5ex]
    \bottomrule[1pt]
    \end{tabular}}
\caption{Top 1 Accuracy on FashionMNIST for different settings of SGLR. 
}  
\label{Results}
\end{table*}
\subsubsection{Impact of SplitAvg}
The SplitAvg algorithm solves the backward client decoupling problem. Using SplitAvg outperforms the baseline where no SplitAvg is used. We provide results for a fraction $\phi$ of 0.25, 0.50 and 0.75. As observed in the Tab. \ref{Results}, using SplitAvg outperforms the baseline. For 20 clients, the accuracy increases by nearly 2\%. Using $\phi=0.75$ performs, in general, the best for the setting with no SplitLr. It is also interesting to note that the increase in the accuracy is more as the number of clients increases when SplitAvg is used. This points towards the potential of SplitAvg when the number of clients is large.


\subsubsection{Integrated Impact of SplitLr \& SplitAvg}
Tab.~\ref{Results} shows the results by combining the two methods of SplitLr and SplitAvg. Both algorithms show scalability individually as well as collectively. For the SplitLr with  $\alpha = 0.5$ SplitAvg with $\phi=0.75$ achieves the highest accuracy in nearly all the experiments. For SplitLr with  $\alpha = 1.0$ the SplitAvg with $\phi=0.75$ again nearly achieves highest accuracy for all experiments. It is interesting to note that for the larger number of clients, $\phi=0.50$ performs on par with $\phi=0.75$ which might indicate that as the number of clients increases, it is better to reduce $\phi$ to keep the number of active clients in a suitable range. 

The accuracy when combining SplitLr and SplitAvg is higher than the set with only one algorithm. This indicates that they can be stacked on top of one another.  To draw a comparison, for 20 clients, without either of the methods, the accuracy is 84.88\%, with SplitAvg($f=0.5$) the top-1 accuracy increases to 86.59\% which is +1.71\%. Further using SplitLr($\alpha=0.5$) the performance increases to 87.48\% which is +0.89\% over the setting of only using SplitAvg. Therefore there is a total increase of 2.60\% by using both SplitLr and SplitAvg.


In the special case, when all the local gradients are averaged, the training is unstable and does not converge as seen in Tab.~\ref{Results Avg Var 1.0}. Using the phased training approach is better but the performance improvement is still lacking. Future work on improving gradient averaging to incorporate more clients will be useful in this respect.
\begin{table}[h]
\centering
    \resizebox{\columnwidth}{!}{
    \begin{tabular}{ccccccc}
    \toprule[1pt]
     \textbf{\emph{C}} & {\color{gray}\textbf{w.o. SplitAvg}} & \multicolumn{5}{c}{\textbf{All SplitAvg}}\\[0.5ex]
      & & \textbf{Initial 60\%} & \textbf{Initial 40\%} & \textbf{Final 60\%} & \textbf{Final 40\%}& \textbf{Full} \\
    \cmidrule(lr){1-1} \cmidrule(lr){2-2} \cmidrule(lr){3-7}
     & & & & & &\\[-1.5ex]
    1 & {\color{gray}82.45} & 82.45 & 82.45 & 82.45 & 82.45 & 82.45 \\
    2 & {\color{gray}81.68} & 81.99 & 81.79 & \textbf{82.21} & 81.47 & 81.82 \\
    4 & {\color{gray}82.93} & \textbf{83.49} & 82.96 & 83.22 & 82.83 & 82.07 \\
    8 & {\color{gray}84.52} & 84.12 & \textbf{84.97} & 84.48 & 83.56 & 83.62 \\
    10 & {\color{gray}84.98} & 84.76 & \textbf{85.32} & 84.47 & 84.31 & 81.61 \\
    15 & {\color{gray}85.59} & 85.12 & 85.20 & 85.05 & \textbf{85.68} & 69.45 \\
    20 & {\color{gray}86.00} & 85.23 & 86.12 & 85.62 & \textbf{86.53} & 60.67\\[0.5ex]
    \bottomrule[1pt]
  \end{tabular}
  }
  \caption{SplitAvg with phased training for SplitLr with $\alpha=1.0$. SplitAvg increases the performance only with phased training but fails to converge without it.}  
  \label{Results Avg Var 1.0}
\end{table}

\subsubsection{Accuracy Comparison with Existing Methods}
We now compare our proposed framework, with other existing works which we had outlined in the Preliminaries section. We present the top 1 accuracy and comput. energy per client in Tab.~\ref{Results Comparison} on the Fashion MNIST dataset \cite{xiao2017fashionmnist} by using AlexNet. The splits of the model are the same for every SL framework where the client model only comprises of 1 convolutional layer and 1 max-pooling layer while the remaining layers are stored in the server. The number of clients has been set to 5 and other hyperparameters have been kept the same for a fair comparison.

Compared to the SL frameworks, as show by Tab.~\ref{Results Comparison}, our method achieves superior accuracy. FL and SGLR are comparable in terms of accuracy to CL which is often considered to be the ceiling. This motivates the question of which method to use. The answer is guided by the communication efficiency and training time which is explored in the Communication Efficiency subsection. 
\begin{table}[h]
\centering
    \resizebox{.7\columnwidth}{!}{
    \begin{tabular}{c c c}
    \toprule[1pt]
    \textbf{Method} & \textbf{Top 1 Acc.} & \textbf{Comput. per Client (PFLOPS)} \\
    \cmidrule(lr){1-1} \cmidrule(lr){2-2} \cmidrule(lr){3-3}
    &\\[-1.5ex]
    CL & 90.5 & 2.944\\
    FL & 89.7 & 0.588\\
    \textbf{SGLR} & 89.3 & 0.031\\
    SFL & 86.0 & 0.031\\
    PSL & 84.7 & 0.031\\[0.5ex]
    \bottomrule[1pt]
    \end{tabular}}
\caption{Top 1 Accuracy on Fashion MNIST dataset with AlexNet. FL and SGLR perform the closest to the CL. FL and CL have large comput. energy cost compared to SL frameworks.}
\label{Results Comparison}
\end{table}
The comput. energy was calculated in peta-FLOPS (PFLOPS) as suggested in OpenAI \cite{openaicompute}. In terms of comput. energy per client, SL frameworks require nearly 94.7\% less energy per client than FL due to the minimal operations carried out at the client side. Therefore CL and FL achieve higher performance at the cost of higher energy consumption.

\subsection{Information Leakage}
Another aspect that we analyse is privacy or information leakage. As described in \cite{geiping2020inverting}, one may extract sensitive information from the intermediate layers to reconstruct and obtain the original data, breaching the privacy of the data set. Therefore we assume a \emph{honest-but-curious} server which attempts to reconstruct the image from the received smashed data after training. We state that SplitAvg creates models whose smashed data are harder to reconstruct the original image with thereby increasing the data privacy.

To validate our claim, we use mutual information as a proxy. For two discrete variables $X$ and $Y$, whose joint probability is given by $\text{P}_{XY}(x,y)$, the mutual information denoted by $I(X;Y)$ can be calculated as
\begin{align}
    I(X;Y) = \Sigma_{x,y}\text{P}_{X,Y}(x,y)\text{log}\frac{\text{P}_{X,Y}(x,y)}{\text{P}_{X}(x)\text{P}_{Y}(y)}
\end{align}
It can also be expressed as the expectation of $\text{log}\frac{\text{P}_{X,Y}(x,y)}{\text{P}_{X}(x)\text{P}_{Y}(y)}$. Similar to \cite{wang2021revisiting}, we utilise the decoder to find the loss between the original image and the reconstructed image formed from the smashed data. We utilise the decoder of a Resnet AutoEncoder which was pretrained on CIFAR-10 from PyTorch Lightning bolts\cite{falcon2020framework}.

We calculate the loss for different variations of SplitAvg by changing the value of $\phi$. 
\begin{table}
\centering
    \resizebox{.8\columnwidth}{!}{
    \begin{tabular}{c c c c c c}
    \toprule[1pt]
    & & & & &\\[-1.5ex]
    \textbf{Fraction $\phi$} & 0.00 & 0.25 & 0.50 & 0.75 & 1.00\\
    \cmidrule(lr){1-1} \cmidrule(lr){2-6}
    \textbf{Loss} & 0.2153 & 0.2142 & 0.2166 & 0.2296 & 0.2759\\
    \bottomrule[1pt]
    \end{tabular}
    }
\caption{Value of cross-entropy loss between the reconstructed image and original image for different values of $\phi$. An increase in the fraction, in general, leads to an increase in the loss indicating the increase in hardness to generate an accurate reconstructed image.}
\label{Privacy Analysis}
\end{table}
The loss values are tabulated in Tab.~\ref{Privacy Analysis}. The results were calculated for 6 clients with 1 client being randomly selected in each case. An increase in $\phi$ leads to an increase in the loss value in general. This demonstrates that the decoder can create a less accurate reconstruction image successively. This points towards less information leakage and an increase in privacy. The initial slow increase can be credited to the small number of clients as with $\phi=0.25$, only 1 client is an active client which practically has no effect. With $\phi=0.5$, 3 clients are active leading to a marginal increase in loss. With $\phi=1.0$ all 6 clients are active when there is quite a large jump in the loss indicating less information leakage. 

\subsection{Communication Efficiency}
We now look at the communication efficiency of the different frameworks. We provide a quantitative analysis of the different frameworks in terms of Training Time and Communication per Epoch in Tab.~\ref{table:Comms Comparison} where the cut layer output size (MB) is denoted as $\text{S}(L)$, the client model size (MB) is FL is denoted by $\text{S}(\mathbf{w})$ and in SL is denoted by $\text{S}(\mathbf{w}_c)$.

In SGLR, the number of data samples per client is $\frac{|\mathcal{D}|}{|\mathcal{C}|}$ and the payload size of the smashed data per client per iteration is $\frac{|\mathcal{D}|\text{S}(L)}{|\mathcal{C}|}$. On the other hand, the gradient payload size per iteration is $(1-\phi)\frac{|\mathcal{D}|\text{S}(L)}{|\mathcal{C}|}$ for the clients that use the non averaged gradients and $\text{S}(L)$ for the clients that use the averaged gradient. The total communicated overhead per iteration \emph{e} per client \emph{i} denoted by $\gamma$ is
\begin{align}
\label{comm sglr per epoch}
\gamma_{i,e} & = \frac{(2-\phi)|\mathcal{D}|\text{S}(L)+\text{S}(L)}{|\mathcal{C}|}
\end{align}
and the communication overhead per epoch for all the client becomes
\begin{align}
\label{Comm SGLR}
\gamma_e & = \Sigma_{i \in \mathcal{C}}\gamma_{i,e} = (2-\phi)|\mathcal{D}|\text{S}(L)+\text{S}(L) 
\end{align}

Analyzing the decrease in total communication overhead of SGLR to SFL for a given setting of 100 clients, $\phi=0.5$, $|\mathcal{D}|=50000 \text{samples}$ and the AlexNet model where $\text{S}(L)=0.024 \text{MB}$ and $\text{S}(\mathbf{w})=200 \text{MB}$ and $\text{S}(\mathbf{w}_c) = 67 \text{MB}$, the percentage reduction in communication of SGLR with respect to SFL is $88.6$\%. Similarly for the case of FL, the percentage decrease is $95.499$\%. There we achieve comparable performance at higher communication efficiency.
\begin{figure}
    \centering
    \includegraphics[width=\columnwidth]{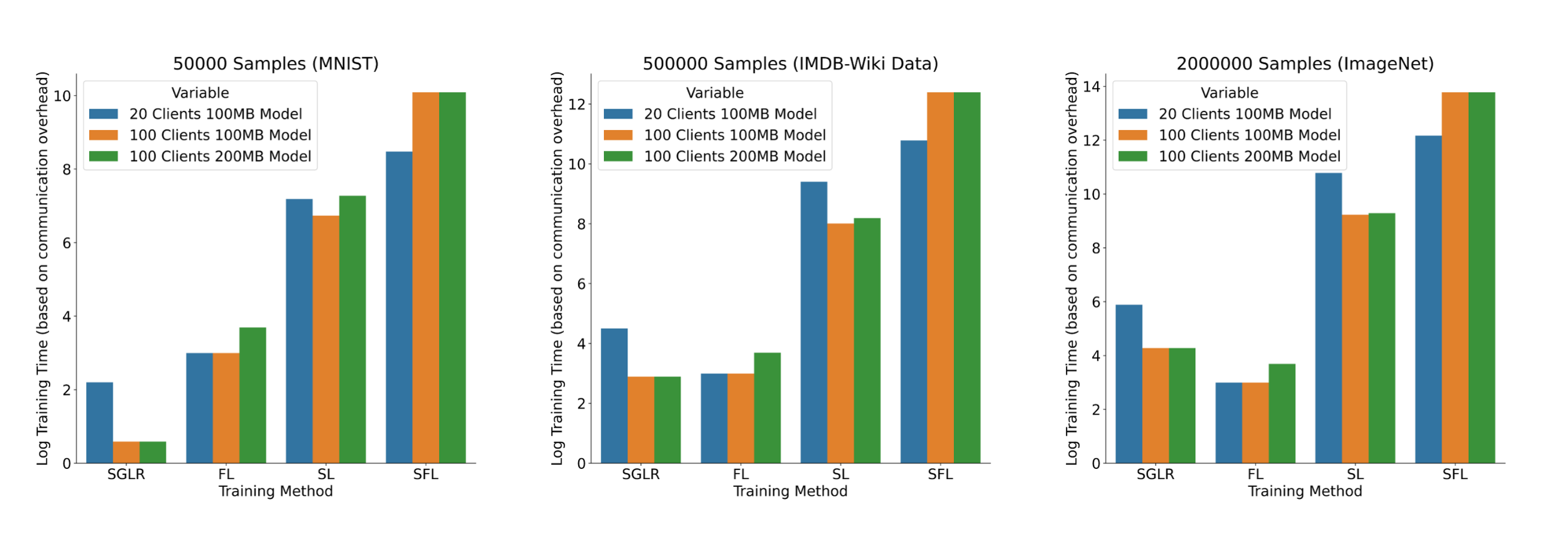}
    \caption{Training Time for SGLR, FL, SL, and SFL for different settings. The plots are for different sizes of the dataset ranging including 50000 (MNIST) , 500000 (IMDB-Wiki) and 2000000 (ImageNet). 3 settings varying model sizes and number of clients for each dataset has been considered for each framework.}
    \label{fig:training time}
\end{figure}
\begin{table}
\centering
    \resizebox{\columnwidth}{!}{
    \begin{tabular}{c c c c}
    \toprule[1pt]
    \textbf{Method} & \textbf{Comm. per Client} & \textbf{Tot. Comm.} & \textbf{Tot. Training Time}\\ 
    \cmidrule(lr){1-1}
    \cmidrule(lr){2-2}
    \cmidrule(lr){3-3}
    \cmidrule(lr){4-4}
    &\\[-1.5ex]
    FL & $2|\text{S}(\mathbf{w})|$ & $2|\mathcal{C}||\text{S}(\mathbf{w})|$ & $T+2\frac{|\text{S}(\mathbf{w})|}{R}$  \\[0.7ex]
    
    SSL & $\frac{2|\mathcal{D}|\text{S}(L)}{|\mathcal{C}|}+2|\text{S}(\mathbf{w}_c)|$ & $2|\mathcal{D}|\text{S}(L)+2|\mathcal{C}||\text{S}(\mathbf{w}_c)|$ & $T+2\frac{|\mathcal{D}|\text{S}(L)}{R}+2|\mathcal{C}|\frac{|\text{S}(\mathbf{w}_{c})|}{R}$\\[0.7ex]
    
    SFL & $\frac{2|\mathcal{D}|\text{S}(L)}{|\mathcal{C}|}+2|\text{S}(\mathbf{w}_c)|$ & $2|\mathcal{D}|\text{S}(L)+2|\mathcal{C}||\text{S}(\mathbf{w}_c)|$ & $T+2\frac{|\mathcal{D}|\text{S}(L)}{|\mathcal{C}|R}+2\frac{|\text{S}(\mathbf{w}_{c})|}{R}$ \\[0.9ex]
    
    \textbf{SGLR} & $\frac{(2-\phi)|\mathcal{D}|\text{S}(L)+\text{S}(L)}{|\mathcal{C}|}$ & $(2-\phi)|\mathcal{D}|\text{S}(L)+\text{S}(L)$ & $T+\frac{(2-\phi)|\mathcal{D}|\text{S}(L)+\text{S}(L)}{|\mathcal{C}|R}$ \\[0.7ex]
    \bottomrule[1pt]
    \end{tabular}
    }
\caption{Communication overhead and total training time analysis of the different training methods of FL, SL, SFL, and SGLR. T is the time required for a forward and backward pass and R is the rate at which the communication occurs. Our method is always faster to train and communication efficient than SL and SFL. It is more efficient than FL when model sizes are large and local data are not too big.}
\label{table:Comms Comparison}
\end{table}

Another important aspect is the training time. Fig.~\ref{fig:training time} extensively demonstrates the training time by varying parameters like the size of the data, the number of clients, and the model size. We assumed $\text{S}(L)$, and the size of client model to be the same as with the analysis of communication efficiency. With large model sizes, SGLR is more efficient than FL where both the methods have comparable accuracy but FL comes out to be more efficient when the total size of the data grows large (ImageNet) though it can be offset when the number of clients is large as well which is usually the case in practical cases when edge devices are considered. 

\section{Conclusion}
In this work, we developed novel learning rate splitting (SplitLr) and split-layer gradient averaging (SplitAvg) algorithms to resolve the two-fold fundamental problem of parallel SL. In the forward propagation, a parallel SL architecture serves multiple clients using a common server model at which the effective batch size increases proportionally with the number of clients, making the server's learning rate slower than desired. In the backward propagation, each client model update is decoupled with other client models, yielding fewer gains from the federation as compared with other scalable distributed learning frameworks such as FL. Motivated by this, we proposed a scalable SL framework SGLR laid by SplitLr and SplitAvg that address such server-side large batch problems and the backward client decoupling problem, respectively.

Simulation results corroborated that SGLR yields less information leakage and higher communication efficiency than the standard parallel SL. This advantage comes mainly from SplitAvg allowing to multicast averaged gradients to multiple clients in common. Combining SplitAvg with SplitLr, SGLR outshines the standard parallel SL in terms of accuracy and scalability for different numbers of clients. The accuracy of SGLR is even on par with FL that may incur larger communication costs and huge information leakage into model-inversion attackers. Extending SplitLr, it is worth developing a learning rate splitting method for cyclic learning rates. Convergence and differential privacy analysis on SplitAvg could also be interesting topics for future work.

\bibliography{aaai22-1}

\end{document}